\documentclass{article}

\usepackage[preprint]{spconf}
\usepackage{array}
\usepackage{booktabs}
\usepackage{cite}
\usepackage{graphicx}
\usepackage{amssymb,amsmath}
\usepackage{framed,multirow}
\usepackage{latexsym}
\usepackage{array}
\usepackage{balance}
\usepackage{booktabs}
\usepackage{caption}
\usepackage{color}
\usepackage{graphicx}
\usepackage{multirow}
\usepackage{times}
\usepackage{url}
\usepackage{threeparttable}
\usepackage{float}

\begin{document}
\title{Progressive Evidence Refinement for Open-domain Multimodal Retrieval Question Answering}
\name{Shuwen Yang$^{\star}$, Anran Wu$^{\star}$, Xingjiao Wu$^{\dagger}$, Luwei Xiao$^{\star}$, Tianlong Ma$^{\star}$, Cheng Jin$^{\dagger}$, Liang He$^{\star}$
\thanks{Shuwen Yang and Anran Wu contributed equally to this work. Corresponding author: Xingjiao Wu (e-mail: xjwu\_cs@fudan.edu.cn).}}

\address{$^{\star}$East China Normal University, Shanghai, China\\
$^{\dagger}$Fudan University, Shanghai, China
\\ \
}
\UseRawInputEncoding
\maketitle

\begin{abstract}

Pre-trained multimodal models have achieved significant success in retrieval-based question answering. However, current multimodal retrieval question-answering models face two main challenges. Firstly, utilizing compressed evidence features as input to the model results in the loss of fine-grained information within the evidence. Secondly, a gap exists between the feature extraction of evidence and the question, which hinders the model from effectively extracting critical features from the evidence based on the given question. We propose a two-stage framework for evidence retrieval and question-answering to alleviate these issues. First and foremost, we propose a progressive evidence refinement strategy for selecting crucial evidence. This strategy employs an iterative evidence retrieval approach to uncover the logical sequence among the evidence pieces. It incorporates two rounds of filtering to optimize the solution space, thus further ensuring temporal efficiency. Subsequently, we introduce a semi-supervised contrastive learning training strategy based on negative samples to expand the scope of the question domain, allowing for a more thorough exploration of latent knowledge within known samples. Finally, in order to mitigate the loss of fine-grained information, we devise a multi-turn retrieval and question-answering strategy to handle multimodal inputs. This strategy involves incorporating multimodal evidence directly into the model as part of the historical dialogue and question. Meanwhile, we leverage a cross-modal attention mechanism to capture the underlying connections between the evidence and the question, and the answer is generated through a decoding generation approach. We validate the model's effectiveness through extensive experiments, achieving outstanding performance on WebQA and MultimodelQA benchmark tests.

\end{abstract}

\begin{keywords}
Web Question Answering, Multimodal Retrival, Transformer.
\end{keywords}

\section{Introduction}
\label{sec:intro}

\begin{figure}[t]
    \centering
    \includegraphics[width=\linewidth]{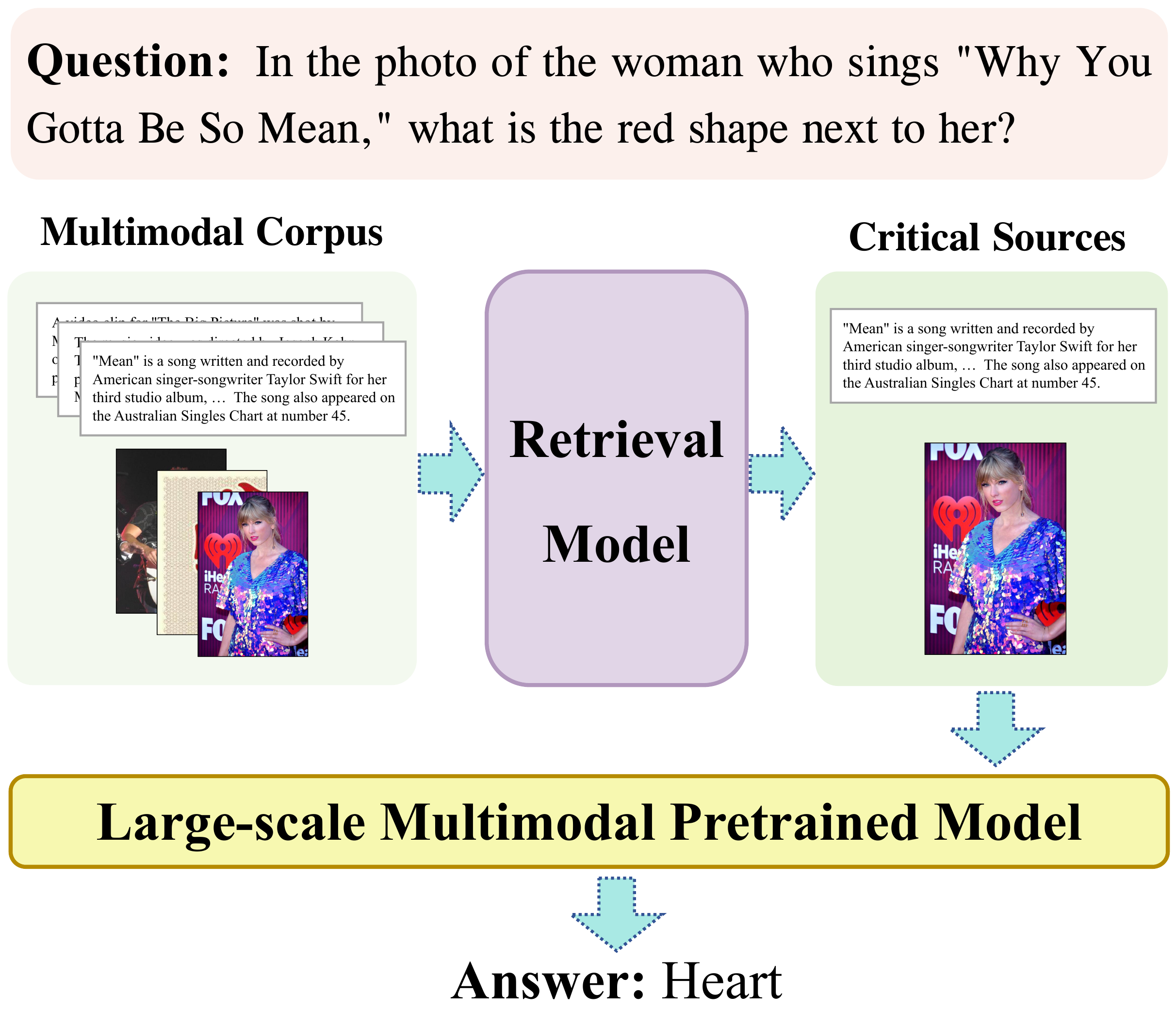}
    \caption{Multimodal Retrieval Question-Answering Overall Workflow. The task requires the system to first identify critical sources from candidate sources through the retrieval system, and then use these key sources as clues to infer the answers to the questions.}
    \label{intro}
\end{figure}

With the continuous advancement of World Wide Web (WWW) technologies, people's activities on the Internet have expanded, resulting in an exponential growth of data generated every day. Due to the sheer volume of information\cite{donratanapat2020national}, individuals struggle to effectively process and sift through this data, leading to the risk of information overload. Furthermore, contemporary internet data is no longer limited to pure text but includes richer multimodal elements such as images, text, and tables. Consequently, extracting crucial information from this vast multimodal information has become a significant challenge\cite{antol2015vqa,rajpurkar2016squad,rajpurkar2018know,schwenk2022okvqa}, enabling better management and utilization of this extensive information resource. Quality, credibility, and accessibility of information have also become focal points to ensure that users can obtain valuable and accurate content from the massive amount of information available. Through persistent efforts, researchers have explored technical approaches using multimodal retrieval question-answering methods to extract key evidence from given multimodal composite data sources and answer questions\cite{chang2022webqa,talmor2020multimodalqa}.

Multimodal retrieval-based question answering involves reasoning by identifying key evidence. However, in practice, the number of candidate evidence is substantial, and these critical pieces of evidence have strict sequential dependencies~\cite{yang2023enhancing}. This leads to exceedingly high time and space complexity in the reasoning process. To effectively address this challenge, existing approaches~\cite{chang2022webqa,chen2022murag,yang2023enhancing} first employ encoders to extract features from the input evidence. Subsequently, pooling operations are utilized to reduce the dimensionality of features. Finally, the reduced-dimensional evidence features are employed for reasoning, and a generation mechanism is employed to generate key evidence and answers. Nevertheless, this approach overlooks specific issues. Firstly, dimensionality reduction results in information loss. Additionally, due to the dimensionality reduction and independently performed feature extraction, a gap exists between the evidence and the original question during feature extraction, leading to a disconnection between the key features extracted and the question.

To mitigate the loss of evidence information, researchers~\cite{gao2022transform,karpukhin2020dense,glass2022re2g} have proposed employing a classification-based approach to traverse the matching of questions and key evidence. 
The specific methodology involves utilizing evidence and question encoders to extract features from the evidence and question. Subsequently, the cosine similarity between the question and the features is calculated or integrated. 
Finally, a classifier is employed to determine whether a correlation exists between individual questions and individual items of evidence. This approach addresses the information loss caused by dimensionality reduction. However, the exhaustive and simplistic classification approach ignores the logical relationships among the evidence.

Through analysis, we have discovered that finding a more optimal approach to maximize the identification of effective evidence and address the curse of dimensionality constitutes a central challenge in multimodal retrieval-based question answering. It is evident that the more correct key evidence we find and the more effectively we connect them, the more accurate the obtained answers will be. Therefore, an intuitive idea arises: Can we eliminate as much irrelevant information as possible? Consequently, we propose a stepwise evidence refinement retrieval strategy comprising two steps: initial screening and iterative retrieval. The initial screening leverages the calculation of distances between question features and candidate features to identify potential candidate evidence while excluding spurious samples that are entirely irrelevant to the question. To enhance the prediction of matching scores between questions and evidence, we employ contrastive learning~\cite{chen2020simple,gao2021simcse} to train the question encoder and evidence encoder. The iterative retrieval step utilizes the question and the evidence identified during the initial screening to further predict key evidence.

However, when applying contrastive learning for initial screening, traditional contrastive learning training often only utilizes key evidence and questions for comparison. The training effectiveness can be severely impacted, especially when there is a significant amount of interfering evidence in the dataset. We propose a negative sample semi-supervised contrastive learning training strategy to overcome this challenge. By reconstructing the question pool, we fully leverage interfering evidence to train the model, effectively filtering out interfering evidence and enhancing the robustness of the initial screening model.

During the question-answering stage, the model needs to reason the answer based on the question and the retrieved key evidence as input. Currently, the general approach for answering questions based on retrieved evidence is directly extracting features from the retrieved multimodal information and then performing question-answering. However, feature extraction results in the loss of some fine-grained details, which hinders effective reasoning based on these details. Therefore, a more favorable approach to address this challenge would be directly inputting the multimodal corpus into the model for question-answering without losing fine-grained information. Inspired by recent works~\cite{li2022mplug,ye2023mplug}, we expect the model to directly process the initial information from the input sources rather than the pooled features of the sources, thereby mitigating the loss of fine-grained information. Thus, we propose a large-scale multimodal processing model that serves the question-answering task with a focus on key evidence. To further ensure the logical coherence among different modalities of evidence, we introduce an innovative multi-turn dialogue mechanism during model training. Through this mechanism, the model can interactively incorporate the evidence as dialogue history with the final input question and capture the intermodal relationships among the evidence using cross-attention mechanisms, allowing multiple images to participate in the answer reasoning process simultaneously.

To validate the efficacy of the proposed method, we conducted extensive experiments on two widely used multi-source visual question-answering datasets WebQA~\cite{chang2022webqa} and MultiModelQA~\cite{talmor2020multimodalqa}. Our approach achieved state-of-the-art (SOTA) performance on both the retrieval and question-answering tasks of the WebQA and MultiModelQA datasets, verifying its effectiveness and highlighting its superiority over existing methods.

In a nutshell, our contributions can be summarized as follows:

\begin{itemize}
\item In order to minimize information loss and fully exploit the logical reasoning relationships among the evidence, we propose a stepwise evidence refinement retrieval strategy. This strategy leverages the inherent logical connections among the evidence to effectively constrain the solution space, thereby enhancing the selection of key evidence. Additionally, our approach also achieves an efficient reduction in space complexity.

\item We propose a negative sample semi-supervised contrastive learning training strategy to address the issue of disregarding interfering samples during the initial screening of contrastive learning. This scheme resolves the problem of underutilization of interfering samples during training, enhancing the robustness of the model and improving the accuracy of the initial screening process.

\item To further ensure the logical coherence among different modalities of evidence, we propose a novel multi-turn dialogue visual question-answering approach for handling the task of visual question answering with multiple-source evidence inputs. This approach mitigates the disconnection between evidence feature extraction and question processing, effectively preserving the latent information among the evidence and significantly boosting question-answering performance.

\item We conduct comprehensive experiments and detailed analysis on publicly available datasets WebQA and MultiModelQA. Our method achieves state-of-the-art (SOTA) results on both the retrieval and question-answering tasks, validating its significance.

\end{itemize}

\section{Related work}
\label{sec:related}

\vspace{2pt}
\noindent
\textbf{Information Retrieval-Based Question and Answering. }
Paragraph retrieval has consistently been a vital component of information retrieval for question-answering tasks\cite{voorhees1999trec}. This task aims to utilize neural networks to locate relevant paragraphs within a corpus, using these paragraphs and questions as context for reasoning and generating answers. Strong sparse vector space models such as TF-IDF or BM25 have traditionally served as standard methods extensively applied to various QA tasks\cite{chen2017reading,yang2019end,nie2019revealing,wolfson2020break}. Recent research has also explored enhancing text-based retrieval by utilizing external structured information, such as knowledge graphs and Wikipedia hyperlinks\cite{min2019knowledge,asai2019learning}.
Existing retrieval-based QA methods can be classified into two categories: cross-attention models and dual-attention models. Cross-attention models\cite{macavaney2019cedr,yang2019end,chen2020cross,zhang2023led,chen2021cross} initially establish connections between queries and documents, then employ a single encoder equipped with cross-attention mechanisms to learn the potential interactions between queries and documents. Typically, cross-attention models\cite{chang2019pre,karpukhin2020dense,khattab2020colbert,qu2021rocketqa,ren2021rocketqav2, santhanam2022colbertv2} can achieve significant retrieval performance improvements but come at a higher computational cost. Dual-attention models, on the other hand, first encode queries and documents separately using distinct encoders and then employ dual-attention mechanisms to facilitate query-document matching.

\vspace{2pt}
\noindent
\textbf{Multimodal Retrieval-Based Question and Answering. }
Multimodal Multi-Hop Question Answering (MMQA) is similar to traditional retrieval-based question-answering tasks\cite{chang2022webqa}, but it differs in that, in MMQA, each question requires retrieving key clues from various multimodal information sources to reason and generate an answer. Additionally, the MMQA task introduces distractor items to evaluate a model's ability to retrieve key evidence and resist interference. Recently, there have been benchmark datasets\cite{chang2022webqa,talmor2020multimodalqa,hannan2020manymodalqa} specifically designed for tackling complex questions that involve multimodal input and contextual reasoning.

Currently, most MMQA methods\cite{gao2022transform,karpukhin2020dense,glass2022re2g} typically encode source data and employ classification layers or maximum inner-product search to select evidence, which is then passed to a decoder for answer generation. However, these methods primarily filter evidence by predicting the relevance between the question and the evidence, making it challenging to identify difficult-to-spot evidence that is less directly related to the question but beneficial for inference. Furthermore, these methods often support single-image input only.
Another category of methods\cite{chang2022webqa,chen2022murag,yang2023enhancing} employs a different strategy by encoding source data into low-dimensional features and feeding these features along with the question to the decoder to predict evidence and generate answers. However, this approach results in significant information loss when compressing evidence, thereby limiting its question-answering performance.

To address these issues, our approach adopts a two-stage evidence retrieval and answer generation process to avoid information loss. Moreover, we design a stepwise refined evidence retrieval strategy to tackle the challenge of identifying difficult key evidence. Additionally, to address the issue of multiple-image input, we introduce a multi-turn dialogue visual-language model to simultaneously process content from multiple images in a multi-turn dialogue manner, thus enhancing the model's performance.

\section{APPROACH}
\label{sec:format}

\begin{figure*}[t]
  \centering
  \includegraphics[width=0.95\linewidth]{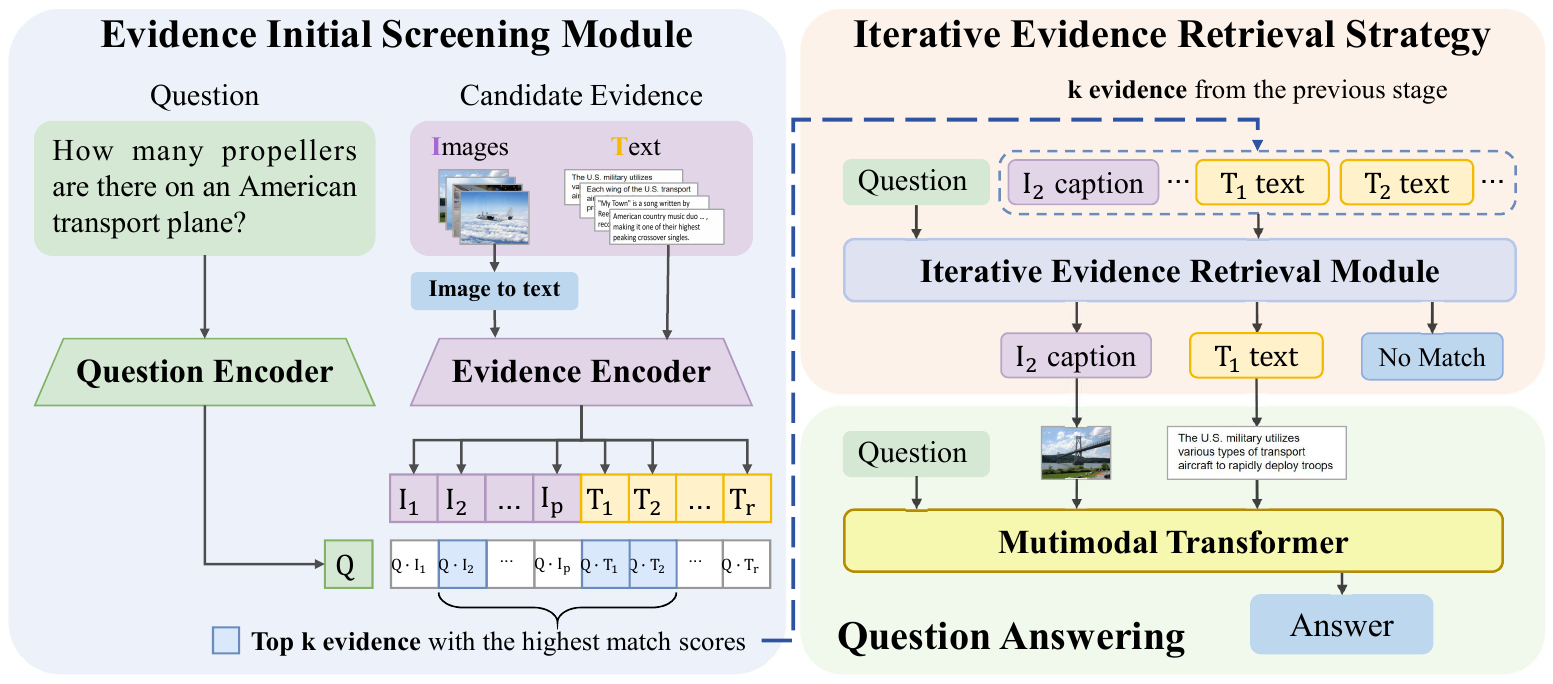}
  \caption{Overall structure diagram of the PERQA model. It is mainly divided into two stages, one is the evidence retrieval stage which gradually refines the evidence, and the other is the question and answer stage based on multi-source input.}
  \label{fig1}
\end{figure*}

This task is divided into two stages. First, given a question $q$ and a set of candidate sources $E = \{{e_0}, {e_1}, ..., {e_n}\}$, where $n$ is the number of candidate sources, which can include images, text, or table information. The task in the first stage is to retrieve the true sources $R = \{{r_0}, {r_1}, ..., {r_m}\}$ from the candidate sources $E$, where $m$ is the number of true sources related to the answer. Then, in the second stage, the answering model needs to generate an answer based on the question $q$ and the selected true sources $R$ as context.

As depicted in Figure \ref{fig1}, our method, namely Progressive Evidence Refinement Question\&Answering (PERQA), consists of two main steps. The first step involves evidence retrieval based on a step-wise evidence refinement strategy, which is used to precisely identify the true sources. The second step is the multi-cue question-answering stage, in which these two stages separately accomplish the tasks of evidence retrieval and answer generation.

\subsection{Evidence Retrieval Stage Based on Progressive Evidence Refinement}

The accuracy of retrieval in the context of a multi-hop retrieval question-answering method directly impacts the final accuracy of the answering process, making this step crucial. This stage is primarily divided into two steps: the first step involves the use of an \textbf{Evidence Initial Screening Module (EISM)} to filter out sources relevant to the question. We employ two simple question and evidence encoders to encode the question and evidence sources separately. We calculate the cosine similarity between the question and candidate source features as a matching score to filter potential genuine sources, thus narrowing down the retrieval scope for the next step. In the second step, we utilize an \textbf{Iterative Evidence Retrieval Strategy (IER)} to determine the true genuine sources from the sources filtered in the previous step.

\subsubsection{Evidence Initial Screening Module (EISM)}

Drawing inspiration from other text-matching methods, the Evidence Initial Screening Module comprises both a question encoder and an evidence encoder, both of which are language transformer models. However, for image sources, that cannot be directly processed by language transformers, we employ an approach that converts images into relevant textual information. In comparison to using CNN\cite{krizhevsky2012imagenet} or ViT\cite{dosovitskiy2020image} directly, this method incurs lower training costs and allows us to overlook the semantic gap between images and text during training. In our work, the image-to-text conversion methods we adopt include image captioning\cite{wang2022ofa} and object detection\cite{girshick2015fast}, summarizing the entire image through overall descriptions and object descriptions. It's worth noting that we exclusively utilize the image-to-text approach during the retrieval stage. In contrast, in the question-answering stage, we directly employ the initial image information to accomplish the answering task.

We assume that the text representation of the $i$-th source is denoted as $S_i$. We start by inputting the question into the question encoder, obtaining the question feature $f_Q$, and then feeding each source into the evidence encoder, yielding source features $f_{S_i}$. Subsequently, we compute the similarity between the question feature and the candidate source features using cosine similarity, which serves as the matching score between the question and the candidate source. The formula is as follows:

\begin{gather}
f_Q = Q_{encoder}(Q) \\
f_{S_i} = E_{encoder}(S_i) \\
P_{cl}(Q, S_i) = \frac{f_Q \cdot f_{S_i}}{\left \| f_Q \right \| \left \| f_{S_i} \right \|}
\end{gather}

where $Q_{encoder}$ represents the question encoder, and $E_{encoder}$ represents the evidence encoder. After calculating the matching scores between the questions and each candidate source, we select the top-k candidate sources with the highest matching scores for further refinement in the next step.

\begin{figure}[t]
    \centering
    \includegraphics[width=0.8\linewidth]{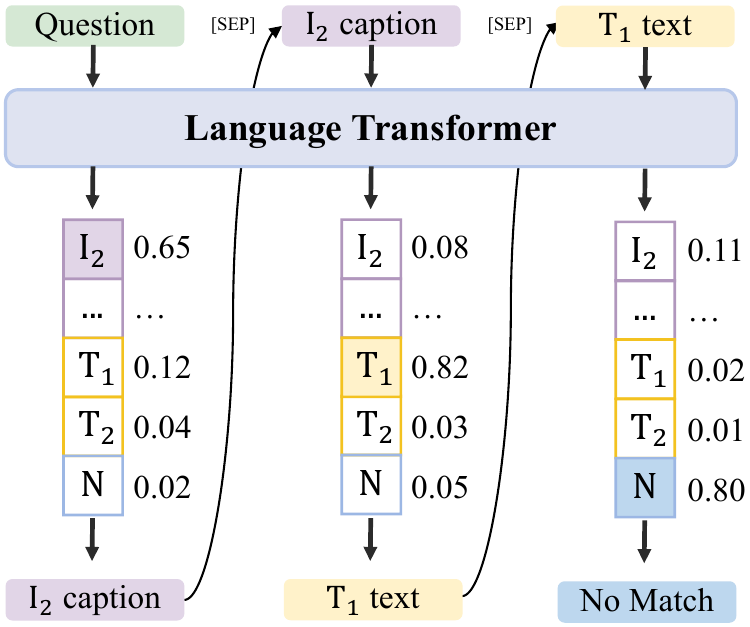}
    \caption{The overall process of Iterative Evidence Retrieval Strategy.}
    \label{fig2}
\end{figure}

\subsubsection{Iterative Evidence Retrieval Strategy (IER)}
In BERT-based question-answering methods\cite{glass2022re2g,nogueira2019passage}, to enhance inference and capture the relevance between questions and paragraphs, it is common to concatenate the question and paragraph and then input them into BERT to predict results. Our approach follows a similar pattern by concatenating the question and evidence and inputting them into BERT to predict their matching scores. This allows us to compute the most relevant evidence for a given question. As each question may have multiple key sources, we can iteratively append sources one after another to the question and the already retrieved sources to determine the matching score for the next source. Using this method, the model can capture implicit relationships between questions and sources, as well as between different sources, uncovering logical ordering relationships among sources, and ultimately enhancing the accuracy of retrieving subsequent sources.

Given a question $Q$ and a set of candidate sources $S$, assuming that we have already determined the set of key sources $R$ and the remaining set of candidate sources $S - R$, we calculate the matching score for the next source using the following method:

\begin{equation}
P_{ie}(Q, R, e) = reg(BERT(Q; R; e))
\end{equation}

\begin{figure*}[t]
    \centering
    \includegraphics[width=1\linewidth]{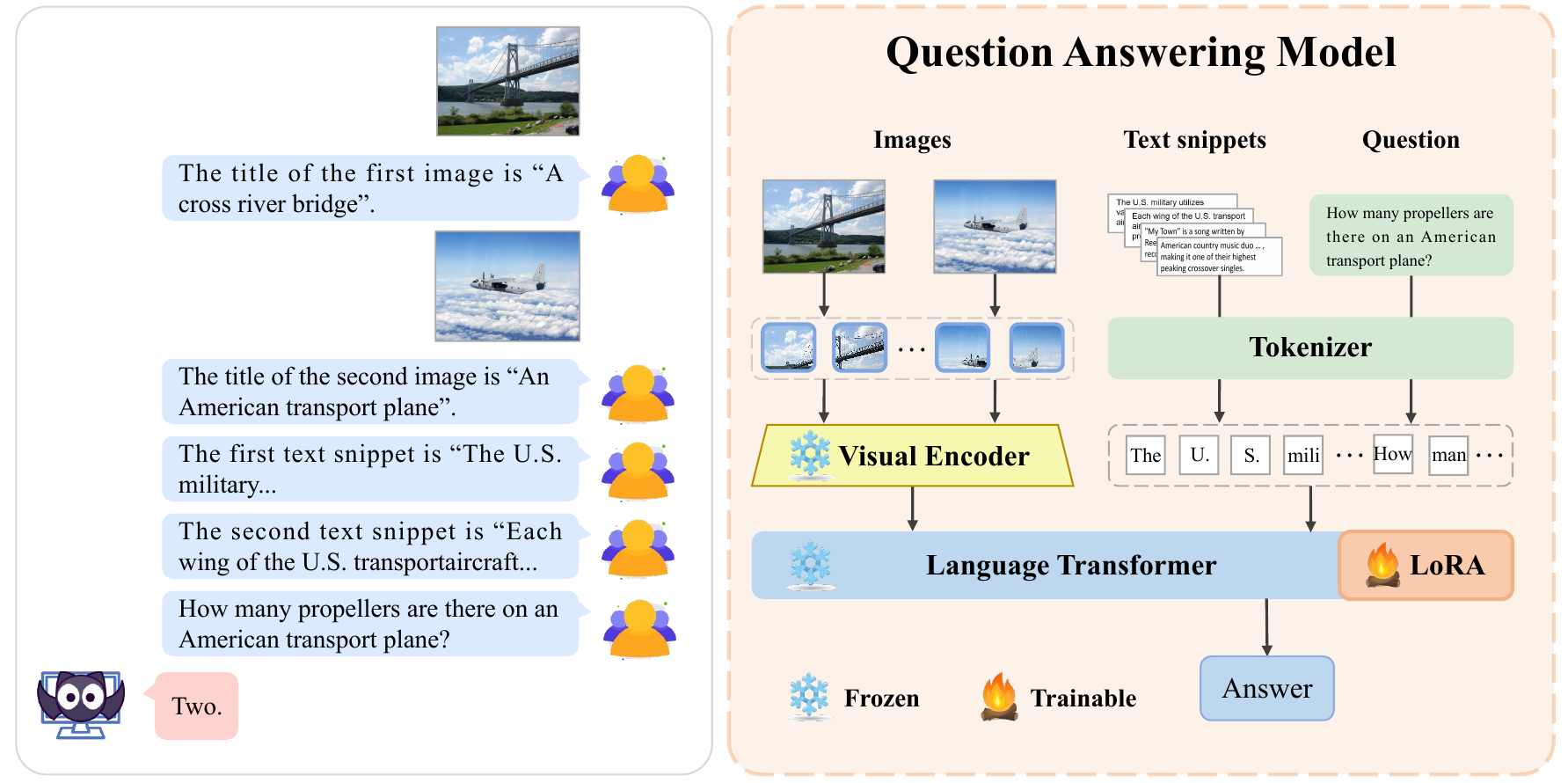}
    \caption{The overall structure of the QA module. Our QA module uses evidence information as a historical dialogue to realize information interaction between evidence.}
    \label{fig3}
\end{figure*}

Here, $reg$ represents a regression module implemented using a linear layer, which is employed to compute the matching score. $e$ is a candidate source such that $e \in S - R$. It is important to note that this process is iterative. Initially, R is an empty set, and S represents the coarse screening results for the first round. Each time a key source is selected, we traverse the set of candidate sources and calculate the matching score. We select the source with the highest score to add to the set of critical sources and initiate the next round of retrieval until a termination signal is reached. In Figure \ref{fig3}, we begin by traversing and selecting the highest-scoring candidate evidence, $I_2$, as the first key evidence. We then concatenate $I_2$'s caption with the question and continue to search for the following key evidence. Similarly, we choose the highest-scoring $T_1$ as the second key evidence, and so on, until we select a termination symbol. Typically, we include the termination symbol in the initial candidate evidence set. Due to the initial round of filtering, the candidate source set is not particularly large, ensuring time efficiency.

\subsection{A question-answering model based on multi-sources input.}

In multi-modal models\cite{li2022mplug,ye2023mplug} that support multi-turn dialogues, images, and text can be linked as a history of conversations by concatenating the features of all images and text. A cross-attention mechanism is then used to capture relationships among these elements. Leveraging this, we utilize such a multimodal model for multi-turn conversations, inputting evidence as part of the conversation history and the question simultaneously. This approach allows us to generate answers. Since we've already acquired the most relevant evidence in the previous stage, we only need to select the most relevant pieces of evidence from the candidate set, significantly reducing computational costs.

As shown in Figure \ref{fig4}, our question-answering model consists of ViT\cite{dosovitskiy2020image} + LLaMA\cite{touvron2023llama} + LORA\cite{hu2021lora} (Low-Rank Adaptation). For a higher-resolution model image, please refer to the appendix. We directly encode image information using ViT to obtain image features and encode text and question features using the LLAMA encoder. In intermediate layers, cross-attention is employed between image features and text features to capture relationships among different pieces of evidence and between evidence and questions. Finally, the LLAMA decoder decodes the combined features to obtain the final answer. Due to the large number of parameters in ViT and LLAMA, we freeze the parameters of the ViT and LLAMA models during training and only fine-tune the parameters of the LORA part.

\subsection{Training}

The training process consists of three steps. First, we train the two encoders used in the first step of evidence retrieval using a semi-supervised contrastive learning method with negative samples. Then, we separately train the iterative evidence retrieval model and the question-answering model using binary cross-entropy and cross-entropy loss functions.

\subsubsection{Negative sample semi-supervised contrastive learning (NSCL)}

During this training phase, we followed the CLIP\cite{radford2021learning} approach, sampling multiple questions from the training set, each accompanied by a key piece of evidence, and organizing them into a batch for training. We use two separate language models to encode the information from both questions and evidence. The similarity between features is computed using cosine similarity as a measure of the match between the questions and evidence.
During training, our goal is to maximize the similarity between the features of each question and its corresponding evidence while minimizing the similarity between the features of other evidence in the same batch. The objective function can be written as follows:

\begin{equation}
    \mathcal{L}_{cl} = -\frac{1}{B}\sum_{i=1}^{B}\log\frac{\mathit{exp}(P_{cl}(Q_i, S_i))}{\sum_{j=1}^{B}\mathit{exp}(P_{cl}(Q_i, S_j))}
\end{equation}

where $B$ represents the batch size, and within the same batch, it contains multiple question-evidence pairs: $\{(Q_0, S_0), (Q_1, S_1), ...,$ $ (Q_B, S_B)\}$. $P_{cl}$ stands for cosine similarity.

It can be observed that the naive contrastive learning training strategy requires each piece of evidence to be paired with a related question, but the dataset does not provide such related questions for distractor evidence. To address this issue, during the training phase, we treat each distractor evidence as its own related question and add it to the batch for training. Formally, for each question $Q_i$, we sample a distractor source $S_{Q_i}^-$ associated with that question and include it in the batch. In this case, a batch of question-evidence pairs is updated as follows: $\{(Q_1, S_1), (S_{Q_1}^-, S_{Q_1}^-), ..., (Q_B, S_B), (S_{Q_B}^-, S_{Q_B}^-)\}$. 

Since the distractor sources for each question may be positively correlated with each other, and contrastive learning does not allow positively correlated samples, we sample only one distractor evidence per question to be added to the batch.

\subsubsection{Iterative Evidence Retrieval Model Training}

During the training phase of the iterative evidence retrieval model, we sample sets of question-key evidence pairs from the dataset. We divide the key evidence set into two parts: the first part serves as already retrieved key evidence, and the remaining key samples are the ones we aim to predict as key evidence, while the distractor evidence serves as negative examples. We want the model to infer the next key evidence based on the question and the already retrieved key evidence, aiming for high matching scores with the remaining key evidence. Formally, assuming the question is $Q$, the retrieved evidence is $R$, the remaining key evidence is $G$, and the distractor evidence set is $N$, we use binary loss as the objective function:

\begin{gather}
\mathcal{L}_{ir} = -(\sum_{e^+ \in G}\log p(e^+) - \sum_{e^- \in N}\log p(e^-)) \\
p(e) = P_{ie}(Q, R, e)
\end{gather}

\subsubsection{Question and Answering Model Training}

For the training of the question and answering model, we employ the commonly used negative log-likelihood as the objective function for text generation. Formally, assuming the question is denoted as Q and the retrieved evidence is denoted as $R$, the objective function for the generation part is defined as:

\begin{gather}
\mathcal{L}_{g} = -\sum_{i=1}^{l}-\log P_g(a_i | R; Q; a_{<i})
\end{gather}

Here, $l$ represents the length of answer tokens, and $P_g$ represents the probability score predicted by the question-answering model for the occurrence of a token.

\section{Experiments}
\subsection{Dataset}

\begin{table*}[t]
\centering
\caption{Overall Statistics of the downstream dataset.}
\begin{tabular}{lccc}
\hline
\multirow{2}{*}{Dataset} & Train            & Dev              & Test             \\
                         & Image/Text/Table & Image/Text/Table & Image/Text/Table \\ \hline
WebQA                    & 18K/17K/0        & 2.5K/2.4K/0      & 3.4K/4K/0        \\
MultimodalQA             & 4.5K/12.6K/7.1K  & 0.8K/1.6K/0.9K   & -                \\ \hline
\end{tabular}
\label{tab1}
\end{table*}

We conducted experiments on two of the most representative MMQA datasets: MultimodalQA\cite{talmor2020multimodalqa} and WebQA\cite{chang2022webqa}. Table \ref{tab1} presents the statistical information for both datasets. 

\noindent \textbf{MultimodalQA\cite{talmor2020multimodalqa}} This dataset comprises multimodal questions from various modalities, manually annotated, including images, text, and tables. The questions in this dataset are generated from templates, with 16 question types and 13 requiring cross-modal retrieval and inference. Since test labels for this dataset have not been released, we report results solely on the validation set. Answers in MultimodalQA typically consist of phrases, and the evaluation metrics employed are Exact Match (EM) and Average F1.

\noindent \textbf{WebQA\cite{chang2022webqa}} This dataset contains multi-hop, multimodal question-answer pairs, where each query requires 1-2 images or 1-2 text snippets to answer. The answers in WebQA are in free-form sentences. There are two evaluation metrics used: one for assessing the model's retrieval accuracy, which is the F1 metric, and another for evaluating the quality of the model's answer generation, which combines QA-FL and QA-Acc based on BARTScore\cite{yuan2021bartscore}. QA-FL measures the fluency (grammatical and semantic coherence) between the generated answer and the reference, while QA-Acc assesses the overlap of key entities between the output answer and the reference.

\subsection{Baselines}
For our comparative experiments, we selected state-of-the-art (SOTA) baseline models from WebQA and MultimodalQA: \textbf{AutoRoute\cite{talmor2020multimodalqa}}, \textbf{ImplicitDec\cite{talmor2020multimodalqa}}, \textbf{VLP\cite{chang2022webqa}}, \textbf{VLP+VinVL\cite{chang2022webqa}}, \textbf{MuRAG\cite{chen2022murag}}, and \textbf{SKURG\cite{yang2023enhancing}}. \textbf{AutoRoute\cite{talmor2020multimodalqa}} identifies the question modality and directs questions and input sources to the corresponding QA modules (textQ, tableQ, or imageQ) by employing a question-type classifier and utilizing different submodels to extract answers. This approach employs RoBERTa-large\cite{liu2019roberta} for question type classification and text-related question answering and uses VILBERT-MT\cite{lu2019vilbert} for visual-related question answering tasks. \textbf{ImplicitDec\cite{talmor2020multimodalqa}} follows a similar method to AutoRoute but introduces a sequential concept. \textbf{VLP\cite{chang2022webqa}} and \textbf{VLP+VinVL\cite{chang2022webqa}} are encoder-decoder models based on transformers. They first extract pooled features of individual modal evidence using pretrained image encoders or text encoders and then concatenate these evidence features with the question as input to predict key evidence and generate answers. \textbf{MuRAG\cite{chen2022murag}} precomputes the encoding of candidate evidence using ViT\cite{dosovitskiy2020image} and BERT\cite{kenton2019bert} and stores it in memory M. It takes a query q as input and retrieves its top-K nearest neighbors from the memory M containing image-text pairs. The retrieval results are then combined with the query q as enhanced input to the main encoder-decoder for answer generation. \textbf{SKURG\cite{yang2023enhancing}} combines evidence features using entity relations as carriers and inputs them into a transformer to generate key evidence and answers. This approach employs Bart\cite{lewis2020bart} and OFA\cite{wang2022ofa} as text and image feature encoders, respectively.

\subsection{Implementation Details}
In the evidence coarse screening stage, both our evidence and question encoding models are based on Bart-base\cite{lewis2020bart}. During training, we utilize AdamW\cite{loshchilov2018decoupled} as the optimizer and employ a linear strategy for learning rate decay. Additionally, in this stage, we set the top-K to 16. To address the scarcity of questions in the dataset, we augment the training set with data from SQuAD\cite{rajpurkar2016squad} and SQuAD2.0\cite{rajpurkar2016squad}. For the image-to-text transformation process, we use OFA-large\cite{wang2022ofa} to extract image descriptions and Fast-RCNN\cite{girshick2015fast} to extract object information from the images.
In the evidence fine screening stage, we employ Deberta-large\cite{he2020deberta} as the backbone. For training, we once again use AdamW as the optimizer. In the question-answering component, we use mPlug-owl as the backbone, and during training, we maintain the use of AdamW as the optimizer. For a comprehensive overview of the experimental configurations, please refer to Table \ref{hyper}. In the table, the first column outlines the experimental settings for the Evidence Initial Screening Module (EISM), the second column for the Iterative Evidence Retrieval Module (IER), and the third column for the Question-Answering Module.

Additionally, it's worth noting that in the ablation experiments, after removing the EISM module, we randomly selected 16 candidates as the initial screening results.

\begin{table}[t]
\centering
\caption{Hyperparameters of the different modules that were used on our method. The EISM means "Evidence Initial Screening Module". IER means the "Iterative Evidence Retrieval Module". Q\&A means the "Question and Answering Module" }
\label{hyper}
\begin{tabular}{lccc}
\hline
              & EISM                                        & IER                                         & Q\&A                  \\ \hline
Batch Size    & 256                                         & 8                                           & 32                    \\
Learning Rate & $2 \times 10^{-4}$ & $2 \times 10^{-5}$ & $10^{-6}$ \\
Optimizer     & AdamW                                       & AdamW                                       & AdamW                 \\
Scheduler     & Linear                                      & Linear                                      & Linear                \\ \hline
\end{tabular}
\end{table}

\begin{table*}[t]
\centering
\caption{Experiment results ($\%$) on WebQA official test-set. The best results are in bold.}
\begin{tabular}{lcccc}
\hline
Model                  & QA-FL $\uparrow$ & QA-ACC $\uparrow$ & QA $\uparrow$ & Retr-F1 $\uparrow$ \\ \hline
VLP(Q-only) {[}2022{]} & 34.9  & 22.2   & 13.4 & -       \\
VLP {[}2022{]}         & 42.6  & 36.7   & 22.6 & 68.9    \\
VLP+VinVL {[}2022{]} & 44.2  & 38.9   & 24.1 & 70.9    \\
MuRAG {[}2022{]}       & 55.7  & 54.6   & 36.1 & 74.6    \\
SKURG {[}2023{]}       & 55.4  & 57.1   & 37.7 & 88.2    \\ \hline
PERQA (ours)           & \textbf{61.7}  & \textbf{63.9}   & \textbf{44.4} & \textbf{89.6}    \\ \hline
\end{tabular}
\label{tab2}
\end{table*}

\begin{table*}[t]
\centering
\caption{Experiment results ($\%$) on MultimodalQA dev-set. The best results are in bold.}
\begin{tabular}{lcccccc}
\hline
\multirow{2}{*}{Model}    & \multicolumn{2}{c}{Multi-modal} & \multicolumn{2}{c}{Single-modal} & \multicolumn{2}{c}{ALL}       \\
                          & EM $\uparrow$     & F1 $\uparrow$     & EM $\uparrow$      & F1 $\uparrow$    & EM $\uparrow$    & F1 $\uparrow$    \\ \hline
AutoRoute {[}2020{]}    & 34.2           & 40.2           & 51.7            & 58.5           & 44.7          & 51.1          \\
ImplicitDecomp {[}2020{]} & 44.6           & 51.2           & 51.6            & 58.4           & 48.8          & 55.5          \\
SKURG {[}2023{]}          & 52.5           & 57.2           & 66.1            & 69.7           & 59.8          & 64.0          \\ \hline
PERQA (ours)              & \textbf{54.7}  & \textbf{60.3}  & \textbf{69.7}   & \textbf{74.1}  & \textbf{62.8} & \textbf{67.8} \\ \hline
\end{tabular}
\label{tab3}
\end{table*}

\begin{table*}[t]
\centering
\caption{Experiment results ($\%$) on subsets of MultimodalQA dev-set for different question types. The best results are in bold.}
\begin{tabular}{lccccc}
\hline
\multirow{2}{*}{Model}   & \multicolumn{2}{c}{Text}      & \multicolumn{2}{c}{Image}     & Text-Image    \\
                         & EM $\uparrow$    & F1 $\uparrow$    & EM $\uparrow$    & F1 $\uparrow$    & EM $\uparrow$    \\ \hline
Question-only {[}2020{]} & 15.4          & 18.4          & 11.0          & 15.6          & 13.8          \\
AutoRoute {[}2020{]}     & 49.5          & 56.9          & 37.8          & 37.8          & 46.6          \\
MuRAG {[}2022{]}         & 60.8          & 67.5          & 58.2          & 58.2          & 60.2          \\
SKURG {[}2023{]}         & 66.7          & 72.7          & 56.1          & 56.1          & 64.2          \\ \hline
PERQA (ours)             & \textbf{74.6} & \textbf{80.5} & \textbf{63.9} & \textbf{64.1} & \textbf{72.0} \\ \hline
\end{tabular}
\label{tab4}
\end{table*}

\subsection{Main Results}
Our results on WebQA\cite{chang2022webqa} are presented in Table \ref{tab2}. "(Q-only)" refers to inference using only the questions. It can be observed that our approach outperforms the previous state-of-the-art (SOTA) methods in all metrics. We surpass SOTA by 1.4\% in the retrieval metric Retr-F1 and by 6.7\% in the crucial QA metric. Additionally, we outperform SOTA by 6.3\% in the QA fluency metric QA-FL and by 6.8\% in the keyword accuracy metric. This demonstrates the robustness of our approach.
Furthermore, it is noteworthy that our approach exhibits a substantial lead in the QA metrics, indicating that using initial evidence information rather than pooled features is more conducive to the accuracy of question answering. This also underscores the effectiveness of multi-turn QA systems in this task. 
In terms of retrieval, our F1 score reaches 89.6\%, which is close to the human-evaluated F1 score of 90.5\%. 
This further validates the effectiveness of our step-by-step evidence retrieval strategy for this task.

Our results on MultiModalQA\cite{talmor2020multimodalqa} are presented in Table \ref{tab3}. "Multi-modal" refers to results for questions requiring multi-modal joint inference, "Single-modal" refers to results for questions requiring single-modal inference, and "ALL" represents results across the entire dataset. It can be observed that our approach outperforms the previous state-of-the-art (SOTA) methods in all metrics, showcasing the stability of our approach.
Furthermore, our approach exhibits significant improvements over SOTA not only in multi-modal inference (EM: 2.2\%, F1: 2.9\%) but also in single-modal inference (EM: 3.6\%, F1: 4.4\%). This demonstrates that our method can enhance the performance of multi-modal retrieval question answering while maintaining high performance in single-modal retrieval question answering.

To perform a fair comparison with MuRAG\cite{chen2022murag} on MultiModalQA, we followed MuRAG's approach by selecting questions of two types, ImageQ and TextQ, from this dataset for testing. The results are shown in Table \ref{tab4}. It is evident that our method significantly outperforms the SOTA models on both of these question types. In comparison to the results on the complete dataset, our performance is even better on ImageQ and TextQ. This is because our model has not been specifically pretrained on TableQ, and its performance on this question type may not be as strong as on ImageQ and TextQ. Moreover, it may not perform as well on the challenging task of multi-modal inference as it does on single-modal question types.

\begin{table*}[t]
    \caption{Ablation Experiments Results ($\%$) on the MultiModalQA Dataset. Here, EISM denotes the Evidence Initial Screening Module, NSCL signifies training the initial screening model using Negative Samples semi-supervised Contrastive Learning, and IER represents the Iterative Evidence Retrieval Module. The best results are in bold.}
    \resizebox{0.98\linewidth}{!}{
    \begin{tabular}{l|l|ccccc|ccccc|ccccc}
    \hline
    \multirow{2}{*}{Row} & \multirow{2}{*}{Model} & \multicolumn{5}{c|}{Multi-modal}                                              & \multicolumn{5}{c|}{Single-modal}                                             & \multicolumn{5}{c}{All}                                                       \\
                         &                        & EM            & F1            & Retr-Pre      & Retr-Rec      & Retr-F1       & EM            & F1            & Retr-Pre      & Retr-Rec      & Retr-F1       & EM            & F1            & Retr-Pre      & Retr-Rec      & Retr-F1       \\ \hline
    1                    & SKURG                  & 52.5          & 57.2          & \textbf{86.1} & 75.7          & 80.6          & 66.1          & 69.7          & 94.7          & 80.2          & 86.7          & 59.8          & 64.0          & \textbf{89.6} & 77.7          & 83.2          \\
    2                    & PERQA (ours)                  & \textbf{54.7} & \textbf{60.3} & 81.7          & \textbf{80.7} & \textbf{81.2} & \textbf{69.7} & \textbf{74.1} & \textbf{95.0} & \textbf{82.7} & \textbf{88.4} & \textbf{62.8} & \textbf{67.8} & 87.1          & \textbf{81.6} & \textbf{84.2} \\
    3                    & - w/o NSCL             & 51.6          & 57.6          & 76.0          & 68.2          & 71.9          & 68.1          & 72.8          & 91.8          & 81.1          & 86.1          & 60.5          & 65.8          & 82.8          & 73.8          & 78.0          \\
    4                    & - w/o EISM             & 45.0          & 50.7          & 68.8          & 60.6          & 64.5          & 54.9          & 57.9          & 61.2          & 53.0          & 56.8          & 50.4          & 54.6          & 65.5          & 57.3          & 61.2          \\
    5                    & - w/o IER              & 33.5          & 38.5          & 62.8          & 47.7          & 54.2          & 61.6          & 66.1          & 62.7          & 69.7          & 66.0          & 48.7          & 53.4          & 62.7          & 57.3          & 59.9          \\
    6                    & - w/o (NSCL + IER)     & 30.4          & 34.8          & 43.3          & 47.3          & 45.2          & 58.4          & 63.5          & 42.4          & 68.0          & 52.3          & 45.6          & 50.3          & 42.9          & 56.3          & 48.7          \\
    7                    & - w/o (EISM + IER)     & 17.7          & 20.6          & 10.7          & 9.5           & 10.1          & 24.7          & 28.4          & 6.1           & 8.3           & 7.0           & 21.5          & 24.8          & 8.2           & 9.0           & 8.6           \\ \hline
    \end{tabular}
    }
    \label{tab6}
    \end{table*}

\begin{table}[t]
\caption{Ablation experiments results ($\%$) on negative sample supervised contrastive learning (NSCL) method. "AUG" denotes the use of additional training data, while "NEG" signifies the utilization of negative samples for semi-supervised training. The best results are in bold.}
\resizebox{\linewidth}{!}{
\begin{tabular}{lcccc}
\hline
\multirow{2}{*}{Model} & \multicolumn{3}{c}{WebQA evaluation set} & \multicolumn{1}{l}{WebQA test set} \\
                       & Recall@8    & Recall@12    & Recall@16   & Retr F1                            \\ \hline
NSCL w/o (AUG+NEG)     & 80.3        & 93.4         & 96.4        & 62.7                               \\
NSCL w/o NEG           & 91.2        & 95.8         & 97.9        & 64.5                               \\
NSCL w/o AUG           & 95.1        & 97.5         & 98.7        & 73.0                               \\ \hline
NSCL                   & \textbf{96.7}        & \textbf{98.7}         & \textbf{99.4}        & \textbf{74.0}                               \\ \hline
\end{tabular}
}
\label{tab5}
\end{table}

\subsection{Ablation Study}
The results from the previous section indicate that PERQA outperforms its counterparts in challenging multi-modal retrieval tasks. In this section, we will further enhance understanding of our method through additional experiments and visualizations.

\vspace{2pt}
\noindent
\textbf{Effect of Negative Samples Semi-Supervised Contrastive Learning. }
We first investigated the impact of NSCL training on the initial screening model. On the validation set of WebQA, we used Recall@k to assess the model's ability to identify key sources from candidate sources. This metric represents the number of key sources identified among the top-k sources with the highest predicted scores divided by the total number of key sources. Finally, we selected the top two sources with the highest predicted scores and a score difference of less than 0.1 to evaluate the overall retrieval ability of this part on the WebQA test set.
The experimental results, as shown in Table \ref{tab5}, indicate the following:
The first row represents the performance of the model trained solely with the naive contrastive learning method for initial screening.
The second row represents the performance after removing the negative sample semi-supervision and instead adding new data to train the model.
The third row shows the performance achieved using only negative sample semi-supervised contrastive learning.
The fourth row represents the performance of the model after full training, including the utilization of the NSCL strategy and additional data during the training process.
By comparing all the data, we can infer that increasing training data and utilizing the negative sample semi-supervised contrastive learning strategy both enhance the model's retrieval capabilities. Furthermore, by comparing the second and third rows, we can see that our proposed negative sample semi-supervised contrastive learning strategy is more effective than simply increasing training samples. However, we also observe that using only initial screening for retrieval does not perform well on the test set. This is because this method is challenging in terms of accurately determining the number of key sources and distinguishing them from indistinguishable distractor sources, making it difficult to improve precision.

\vspace{2pt}
\noindent
\textbf{Effect of Model Components on Overall Performance. }
We conducted a series of experiments on the MultimodalQA dataset to investigate the impact of different model components on overall performance. We used EM and F1 scores provided by the dataset as evaluation metrics for the question-answering task and Retr-Pre, Retr-Rec, and Retr-F1 for evidence retrieval. Firstly, we compared our model with the existing state-of-the-art (SOTA) model SKURG\cite{yang2023enhancing} in MultiModalQA. The results showed that our approach outperformed the SOTA model in both the question-answering and retrieval tasks. It's worth noting that our model achieved slightly lower Retr-Pre scores than SKURG in the multi-modal evidence retrieval task but surpassed SKURG in terms of Retr-Rec and Retr-F1 scores. Our analysis suggests that this is because our model sets lower standards when selecting key sources, allowing it to extract more key sources and thus improving recall. However, this approach also introduces more noise sources, leading to lower precision. Nevertheless, we believe that the model's performance in recall is more critical, as question-answering models can be trained to mitigate interference from noise sources, but without key sources, the model cannot correctly answer questions.

\begin{figure}[t]
    \centering
    \includegraphics[width=1\linewidth]{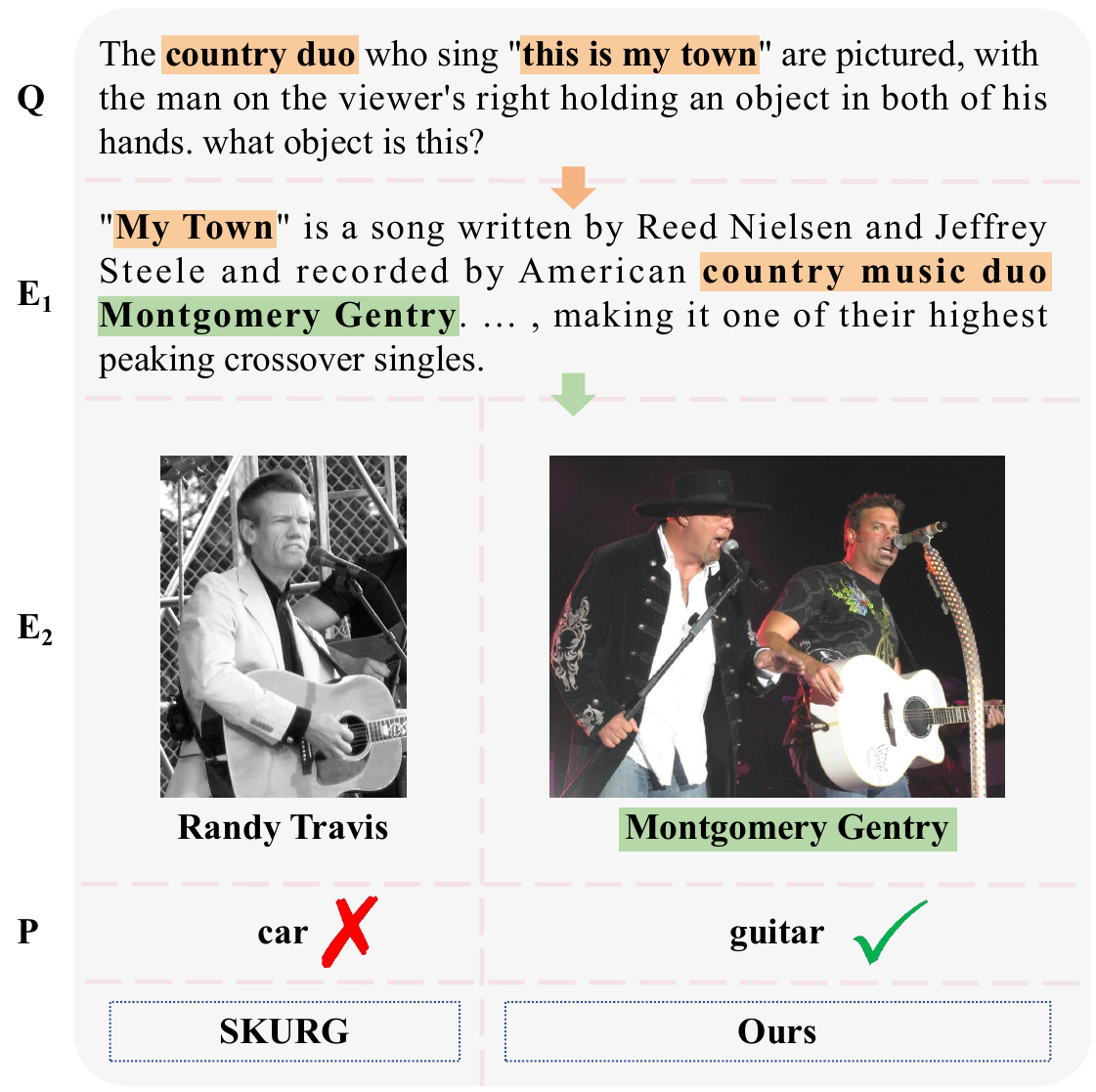}
    \caption{Examples of retrieval Q\&A. Q represents the question, $E_1$ represents the first evidence, $E_2$ represents the second evidence, and P represents the model prediction result.}
    \label{fig4}
\end{figure}

The third row (w/o NSCL) signifies that we retained the initial screening module but did not train the model with NSCL. When compared to the second row, the model's performance in both single-modal and multi-modal retrieval and question-answering accuracy decreased, indicating that NSCL training is essential for this framework. The data in the fourth and fifth rows represent the model's accuracy after removing the evidence initial screening module and the iterative retrieval module, respectively. We can observe a significant drop in recall accuracy after removing EISM, demonstrating that the initial screening module not only effectively reduces the retrieval space for iterative evidence retrieval but also enhances overall retrieval correctness. By comparing the data in the fourth and fifth rows, we can also see that, compared to removing the EISM module, removing the IER module results in worse overall performance in multi-modal scenarios (EM decreased by 11.5\%, Retr-F1 decreased by 10.3\%), but better performance in single-modal scenarios (EM improved by 6.7\%, Retr-F1 improved by 9.0\%). This indicates that the IER module excels in multi-modal multi-hop retrieval, while the EISM module performs better in single-modal retrieval. Combining both modules allows for effective complementarity. The sixth row represents our removal of the IER module, retaining only the initial screening module, and training the model with regular contrastive learning, while the seventh row represents the complete removal of all screening stages and random selection of key evidence from candidate sources for question-answering. Compared to the second row, there is a significant decline in model performance, highlighting the outstanding performance of our proposed method. The question-answering performance of the model that randomly selects key evidence from candidate sources is also less than ideal, emphasizing the necessity of the retrieval stage. Through this series of ablation experiments, we further demonstrate the effectiveness of our proposed method.

\subsection{Case Study}

To demonstrate that our model can complete retrieval and answer inference by mining logical order relationships among pieces of evidence, we have provided an illustrative example. As shown in Figure \ref{fig4}, we present the retrieval and inference process of our method (on the right) and the state-of-the-art (SOTA) SKURG\cite{yang2023enhancing} method (on the left) for a specific question. In this figure, $E_1$ represents intermediate key evidence, which is associated with the question through the orange keywords and linked to other key evidence through the green keywords. $E_2$ is the final key evidence, crucial for reasoning towards the answer. The model needs to correctly retrieve $E_2$ to answer the question. We can observe that both our method and SKURG are able to retrieve the intermediate key evidence $E_1$. However, SKURG incorrectly estimates the final key evidence $E_2$, while our method successfully retrieves it, leading to a correct answer in our case and an incorrect one in SKURG's case. This example illustrates how our method leverages potential logical order relationships among key evidence to retrieve them and answer questions accurately.

\section{Conclusion}
\label{sec:conclusion}

In this work, we propose a two-stage evidence retrieval question-answering method. Our proposed method, named PERQA, utilizes a progressive evidence refinement strategy for evidence retrieval and employs an iterative evidence retrieval approach to elucidate the logical sequence among pieces of evidence. Additionally, we introduce a semi-supervised training strategy based on negative samples to fully exploit latent information in known samples within our retrieval model. Finally, we design a multi-turn question-answering strategy to handle multi-modal inputs, incorporating multiple key pieces of evidence as historical context along with the question input into the model and leveraging cross-modal attention mechanisms to capture potential connections between evidence and questions. Through extensive experiments, we demonstrate the outstanding performance of our method in multi-modal retrieval question-answering tasks, and we look forward to providing valuable insights for future research.

\bibliographystyle{IEEEbib}
\bibliography{total}

\begin{thebibliography}{10}

\bibitem{donratanapat2020national}
N~Donratanapat, S~Samadi, Jos{\'e}~M Vidal, and S~Sadeghi Tabas,
\newblock ``A national scale big data analytics pipeline to assess the potential impacts of flooding on critical infrastructures and communities,''
\newblock {\em Environmental Modelling \& Software}, vol. 133, pp. 104828, 2020.

\bibitem{antol2015vqa}
Stanislaw Antol, Aishwarya Agrawal, Jiasen Lu, Margaret Mitchell, Dhruv Batra, C~Lawrence Zitnick, and Devi Parikh,
\newblock ``Vqa: Visual question answering,''
\newblock in {\em International Conference on Computer Vision (ICCV)}, 2015, pp. 2425--2433.

\bibitem{rajpurkar2016squad}
Pranav Rajpurkar, Jian Zhang, Konstantin Lopyrev, and Percy Liang,
\newblock ``Squad: 100,000+ questions for machine comprehension of text,''
\newblock in {\em Conference on Empirical Methods in Natural Language Processing (EMNLP)}, 2016, pp. 2383--2392.

\bibitem{rajpurkar2018know}
Pranav Rajpurkar, Robin Jia, and Percy Liang,
\newblock ``Know what you don’t know: Unanswerable questions for squad,''
\newblock in {\em Proceedings of the 56th Annual Meeting of the Association for Computational Linguistics (Volume 2: Short Papers)}, 2018, pp. 784--789.

\bibitem{schwenk2022okvqa}
Dustin Schwenk, Apoorv Khandelwal, Christopher Clark, Kenneth Marino, and Roozbeh Mottaghi,
\newblock ``A-okvqa: A benchmark for visual question answering using world knowledge,''
\newblock in {\em European Conference on Computer Vision (ECCV)}. Springer, 2022, pp. 146--162.

\bibitem{chang2022webqa}
Yingshan Chang, Mridu Narang, Hisami Suzuki, Guihong Cao, Jianfeng Gao, and Yonatan Bisk,
\newblock ``Webqa: Multihop and multimodal qa,''
\newblock in {\em Conference on Computer Vision and Pattern Recognition (CVPR)}, 2022, pp. 16495--16504.

\bibitem{talmor2020multimodalqa}
Alon Talmor, Ori Yoran, Amnon Catav, Dan Lahav, Yizhong Wang, Akari Asai, Gabriel Ilharco, Hannaneh Hajishirzi, and Jonathan Berant,
\newblock ``Multimodalqa: complex question answering over text, tables and images,''
\newblock in {\em International Conference on Learning Representations (ICLR)}, 2020.

\bibitem{yang2023enhancing}
Qian Yang, Qian Chen, Wen Wang, Baotian Hu, and Min Zhang,
\newblock ``Enhancing multi-modal and multi-hop question answering via structured knowledge and unified retrieval-generation,''
\newblock {\em ACM International Conference on Multimedia (ACMMM)}, 2023.

\bibitem{chen2022murag}
Wenhu Chen, Hexiang Hu, Xi~Chen, Pat Verga, and William Cohen,
\newblock ``Murag: Multimodal retrieval-augmented generator for open question answering over images and text,''
\newblock in {\em Conference on Empirical Methods in Natural Language Processing (EMNLP)}, 2022, pp. 5558--5570.

\bibitem{gao2022transform}
Feng Gao, Qing Ping, Govind Thattai, Aishwarya Reganti, Ying~Nian Wu, and Prem Natarajan,
\newblock ``Transform-retrieve-generate: Natural language-centric outside-knowledge visual question answering,''
\newblock in {\em Conference on Computer Vision and Pattern Recognition (CVPR)}, 2022, pp. 5067--5077.

\bibitem{karpukhin2020dense}
Vladimir Karpukhin, Barlas Oguz, Sewon Min, Patrick Lewis, Ledell Wu, Sergey Edunov, Danqi Chen, and Wen-tau Yih,
\newblock ``Dense passage retrieval for open-domain question answering,''
\newblock in {\em Conference on Empirical Methods in Natural Language Processing (EMNLP)}. Association for Computational Linguistics, 2020.

\bibitem{glass2022re2g}
Michael Glass, Gaetano Rossiello, Md~Faisal~Mahbub Chowdhury, Ankita~Rajaram Naik, Pengshan Cai, and Alfio Gliozzo,
\newblock ``Re2g: Retrieve, rerank, generate,''
\newblock in {\em Annual Conference of the North American Chapter of the Association for Computational Linguistics (NAACL)}, 2022.

\bibitem{chen2020simple}
Ting Chen, Simon Kornblith, Mohammad Norouzi, and Geoffrey Hinton,
\newblock ``A simple framework for contrastive learning of visual representations,''
\newblock in {\em Proceedings of International Conference on Machine Learning (ICML)}. PMLR, 2020, pp. 1597--1607.

\bibitem{gao2021simcse}
Tianyu Gao, Xingcheng Yao, and Danqi Chen,
\newblock ``Simcse: Simple contrastive learning of sentence embeddings,''
\newblock in {\em Conference on Empirical Methods in Natural Language Processing (EMNLP)}, 2021, pp. 6894--6910.

\bibitem{li2022mplug}
Chenliang Li, Haiyang Xu, Junfeng Tian, Wei Wang, Ming Yan, Bin Bi, Jiabo Ye, He~Chen, Guohai Xu, Zheng Cao, et~al.,
\newblock ``mplug: Effective and efficient vision-language learning by cross-modal skip-connections,''
\newblock in {\em Conference on Empirical Methods in Natural Language Processing (EMNLP)}, 2022, pp. 7241--7259.

\bibitem{ye2023mplug}
Qinghao Ye, Haiyang Xu, Guohai Xu, Jiabo Ye, Ming Yan, Yiyang Zhou, Junyang Wang, Anwen Hu, Pengcheng Shi, Yaya Shi, et~al.,
\newblock ``mplug-owl: Modularization empowers large language models with multimodality,''
\newblock {\em arXiv preprint arXiv:2304.14178}, 2023.

\bibitem{voorhees1999trec}
E~VOORHEES,
\newblock ``The trec-8 question answering track report,''
\newblock in {\em Proceedings of the Text Retrieval Conference (TREC)}, 1999.

\bibitem{chen2017reading}
Danqi Chen, Adam Fisch, Jason Weston, and Antoine Bordes,
\newblock ``Reading wikipedia to answer open-domain questions,''
\newblock in {\em Annual Meeting of the Association for Computational Linguistics (ACL)}, 2017, pp. 1870--1879.

\bibitem{yang2019end}
Wei Yang, Yuqing Xie, Aileen Lin, Xingyu Li, Luchen Tan, Kun Xiong, Ming Li, and Jimmy Lin,
\newblock ``End-to-end open-domain question answering with bertserini,''
\newblock in {\em Annual Conference of the North American Chapter of the Association for Computational Linguistics (NAACL)}, 2019, pp. 72--77.

\bibitem{nie2019revealing}
Yixin Nie, Songhe Wang, and Mohit Bansal,
\newblock ``Revealing the importance of semantic retrieval for machine reading at scale,''
\newblock in {\em Conference on Empirical Methods in Natural Language Processing (EMNLP)}, 2019, pp. 2553--2566.

\bibitem{wolfson2020break}
Tomer Wolfson, Mor Geva, Ankit Gupta, Matt Gardner, Yoav Goldberg, Daniel Deutch, and Jonathan Berant,
\newblock ``Break it down: A question understanding benchmark,''
\newblock {\em Transactions of the Association for Computational Linguistics (TACL)}, vol. 8, pp. 183--198, 2020.

\bibitem{min2019knowledge}
Sewon Min, Danqi Chen, Luke Zettlemoyer, and Hannaneh Hajishirzi,
\newblock ``Knowledge guided text retrieval and reading for open domain question answering,''
\newblock {\em arXiv preprint arXiv:1911.03868}, 2019.

\bibitem{asai2019learning}
Akari Asai, Kazuma Hashimoto, Hannaneh Hajishirzi, Richard Socher, and Caiming Xiong,
\newblock ``Learning to retrieve reasoning paths over wikipedia graph for question answering,''
\newblock in {\em International Conference on Learning Representations (ICLR)}, 2019.

\bibitem{macavaney2019cedr}
Sean MacAvaney, Andrew Yates, Arman Cohan, and Nazli Goharian,
\newblock ``Cedr: Contextualized embeddings for document ranking,''
\newblock in {\em Annual ACM Conference on Research and Development in Information Retrieval}, 2019, pp. 1101--1104.

\bibitem{chen2020cross}
Dongmei Chen, Sheng Zhang, Xin Zhang, and Kaijing Yang,
\newblock ``Cross-lingual passage re-ranking with alignment augmented multilingual bert,''
\newblock {\em IEEE Access}, vol. 8, pp. 213232--213243, 2020.

\bibitem{zhang2023led}
Kai Zhang, Chongyang Tao, Tao Shen, Can Xu, Xiubo Geng, Binxing Jiao, and Daxin Jiang,
\newblock ``Led: Lexicon-enlightened dense retriever for large-scale retrieval,''
\newblock in {\em Proceedings of the ACM Web Conference (ACMWeb)}, 2023, pp. 3203--3213.

\bibitem{chen2021cross}
Cen Chen, Chengyu Wang, Minghui Qiu, Dehong Gao, Linbo Jin, and Wang Li,
\newblock ``Cross-domain knowledge distillation for retrieval-based question answering systems,''
\newblock in {\em Proceedings of the ACM Web Conference (ACMWeb)}, 2021, pp. 2613--2623.

\bibitem{chang2019pre}
Wei-Cheng Chang, X~Yu Felix, Yin-Wen Chang, Yiming Yang, and Sanjiv Kumar,
\newblock ``Pre-training tasks for embedding-based large-scale retrieval,''
\newblock in {\em International Conference on Learning Representations (ICLR)}, 2019.

\bibitem{khattab2020colbert}
Omar Khattab and Matei Zaharia,
\newblock ``Colbert: Efficient and effective passage search via contextualized late interaction over bert,''
\newblock in {\em Annual ACM Conference on Research and Development in Information Retrieval}, 2020, pp. 39--48.

\bibitem{qu2021rocketqa}
Yingqi Qu, Yuchen Ding, Jing Liu, Kai Liu, Ruiyang Ren, Wayne~Xin Zhao, Daxiang Dong, Hua Wu, and Haifeng Wang,
\newblock ``Rocketqa: An optimized training approach to dense passage retrieval for open-domain question answering,''
\newblock in {\em Annual Conference of the North American Chapter of the Association for Computational Linguistics (NAACL)}, 2021, pp. 5835--5847.

\bibitem{ren2021rocketqav2}
Ruiyang Ren, Yingqi Qu, Jing Liu, Wayne~Xin Zhao, Qiaoqiao She, Hua Wu, Haifeng Wang, and Ji-Rong Wen,
\newblock ``Rocketqav2: A joint training method for dense passage retrieval and passage re-ranking,''
\newblock in {\em Conference on Empirical Methods in Natural Language Processing (EMNLP)}, 2021, pp. 2825--2835.

\bibitem{santhanam2022colbertv2}
Keshav Santhanam, Omar Khattab, Jon Saad-Falcon, Christopher Potts, and Matei Zaharia,
\newblock ``Colbertv2: Effective and efficient retrieval via lightweight late interaction,''
\newblock in {\em Annual Conference of the North American Chapter of the Association for Computational Linguistics (NAACL)}, 2022, pp. 3715--3734.

\bibitem{hannan2020manymodalqa}
Darryl Hannan, Akshay Jain, and Mohit Bansal,
\newblock ``Manymodalqa: Modality disambiguation and qa over diverse inputs,''
\newblock in {\em The AAAI Conference on Artificial Intelligence (AAAI)}, 2020, vol.~34, pp. 7879--7886.

\bibitem{krizhevsky2012imagenet}
Alex Krizhevsky, Ilya Sutskever, and Geoffrey~E Hinton,
\newblock ``Imagenet classification with deep convolutional neural networks,''
\newblock {\em Advances in Neural Information Processing Systems (NeurIPS)}, vol. 25, 2012.

\bibitem{dosovitskiy2020image}
Alexey Dosovitskiy, Lucas Beyer, Alexander Kolesnikov, Dirk Weissenborn, Xiaohua Zhai, Thomas Unterthiner, Mostafa Dehghani, Matthias Minderer, Georg Heigold, Sylvain Gelly, et~al.,
\newblock ``An image is worth 16x16 words: Transformers for image recognition at scale,''
\newblock in {\em International Conference on Learning Representations (ICLR)}, 2020.

\bibitem{wang2022ofa}
Peng Wang, An~Yang, Rui Men, Junyang Lin, Shuai Bai, Zhikang Li, Jianxin Ma, Chang Zhou, Jingren Zhou, and Hongxia Yang,
\newblock ``Ofa: Unifying architectures, tasks, and modalities through a simple sequence-to-sequence learning framework,''
\newblock in {\em Proceedings of International Conference on Machine Learning (ICML)}. PMLR, 2022, pp. 23318--23340.

\bibitem{girshick2015fast}
Ross Girshick,
\newblock ``Fast r-cnn,''
\newblock in {\em International Conference on Computer Vision (ICCV)}, 2015, pp. 1440--1448.

\bibitem{nogueira2019passage}
Rodrigo Nogueira and Kyunghyun Cho,
\newblock ``Passage re-ranking with bert,''
\newblock {\em arXiv preprint arXiv:1901.04085}, 2019.

\bibitem{touvron2023llama}
Hugo Touvron, Thibaut Lavril, Gautier Izacard, Xavier Martinet, Marie-Anne Lachaux, Timoth{\'e}e Lacroix, Baptiste Rozi{\`e}re, Naman Goyal, Eric Hambro, Faisal Azhar, et~al.,
\newblock ``Llama: Open and efficient foundation language models,''
\newblock {\em arXiv preprint arXiv:2302.13971}, 2023.

\bibitem{hu2021lora}
Edward~J Hu, Phillip Wallis, Zeyuan Allen-Zhu, Yuanzhi Li, Shean Wang, Lu~Wang, Weizhu Chen, et~al.,
\newblock ``Lora: Low-rank adaptation of large language models,''
\newblock in {\em International Conference on Learning Representations (ICLR)}, 2021.

\bibitem{radford2021learning}
Alec Radford, Jong~Wook Kim, Chris Hallacy, Aditya Ramesh, Gabriel Goh, Sandhini Agarwal, Girish Sastry, Amanda Askell, Pamela Mishkin, Jack Clark, et~al.,
\newblock ``Learning transferable visual models from natural language supervision,''
\newblock in {\em Proceedings of International Conference on Machine Learning (ICML)}. PMLR, 2021, pp. 8748--8763.

\bibitem{yuan2021bartscore}
Weizhe Yuan, Graham Neubig, and Pengfei Liu,
\newblock ``Bartscore: Evaluating generated text as text generation,''
\newblock {\em Advances in Neural Information Processing Systems (NeurIPS)}, vol. 34, pp. 27263--27277, 2021.

\bibitem{liu2019roberta}
Yinhan Liu, Myle Ott, Naman Goyal, Jingfei Du, Mandar Joshi, Danqi Chen, Omer Levy, Mike Lewis, Luke Zettlemoyer, and Veselin Stoyanov,
\newblock ``Roberta: A robustly optimized bert pretraining approach,''
\newblock {\em arXiv preprint arXiv:1907.11692}, 2019.

\bibitem{lu2019vilbert}
Jiasen Lu, Dhruv Batra, Devi Parikh, and Stefan Lee,
\newblock ``Vilbert: Pretraining task-agnostic visiolinguistic representations for vision-and-language tasks,''
\newblock {\em Advances in Neural Information Processing Systems (NeurIPS)}, vol. 32, 2019.

\bibitem{kenton2019bert}
Jacob Devlin Ming-Wei~Chang Kenton and Lee~Kristina Toutanova,
\newblock ``Bert: Pre-training of deep bidirectional transformers for language understanding,''
\newblock in {\em Annual Conference of the North American Chapter of the Association for Computational Linguistics (NAACL)}, 2019, pp. 4171--4186.

\bibitem{lewis2020bart}
Mike Lewis, Yinhan Liu, Naman Goyal, Marjan Ghazvininejad, Abdelrahman Mohamed, Omer Levy, Veselin Stoyanov, and Luke Zettlemoyer,
\newblock ``Bart: Denoising sequence-to-sequence pre-training for natural language generation, translation, and comprehension,''
\newblock in {\em Annual Meeting of the Association for Computational Linguistics (ACL)}, 2020, pp. 7871--7880.

\bibitem{loshchilov2018decoupled}
Ilya Loshchilov and Frank Hutter,
\newblock ``Decoupled weight decay regularization,''
\newblock in {\em International Conference on Learning Representations (ICLR)}, 2018.

\bibitem{he2020deberta}
Pengcheng He, Xiaodong Liu, Jianfeng Gao, and Weizhu Chen,
\newblock ``Deberta: Decoding-enhanced bert with disentangled attention,''
\newblock in {\em International Conference on Learning Representations (ICLR)}, 2020.

\end{thebibliography}

\end{document}